\documentclass[a4paper,fleqn]{cas-sc}

\usepackage[authoryear]{natbib}

\usepackage{siunitx}
\usepackage{calrsfs}
\usepackage[frozencache=true, cachedir=minted-cache]{minted}
\usepackage{algpseudocode}
\usepackage{algorithm}
\usepackage{algorithmicx}
\usepackage{algpseudocode}
\usepackage{listings}
\usepackage{array}
\newcommand{\Lmo}[0]{L_\mathrm{mo}}

\newcommand{\concat}[0]{\bigoplus}
\newcommand{\R}[0]{\mathbb{R}}

\newcommand{\ugnd}[0]{u_\mathrm{gnd\text{-}sn}}
\newcommand{\uobs}[0]{u_\mathrm{obs\text{-}sn}}
\newcommand{\ugeom}[0]{u_\mathrm{geom\text{-}sn}}
\newcommand{\uvol}[0]{u_\mathrm{vol}}

\algblock{Input}{EndInput}
\algnotext{EndInput}
\algblock{Output}{EndOutput}
\algnotext{EndOutput}
\algblock{Code}{EndCode}
\algnotext{EndCode}

\begin{document}
\let\WriteBookmarks\relax
\def\floatpagepagefraction{1}
\def\textpagefraction{.001}

\shorttitle{Anchored-Branched Steady-state WInd Flow Transformer}

\shortauthors{A. d. Villeroché et.~al}

\title [mode = title]{Anchored-Branched Steady-state WInd Flow Transformer (AB-SWIFT): a metamodel for 3D atmospheric flow in urban environments}  

\author[1]{Armand de Villeroché}[
      orcid=0009-0001-8811-7443,]\corref{cor1}
\ead{armand.de-villeroche@edf.fr}
\credit{Conceptualization, Methodology, Software, Validation, Data Curation, Writing - Original Draft}
\author[1,3]{Rem\textendash Sophia Mouradi}
\credit{Supervision, Writing - Review \& Editing}
\author[2,3]{Vincent Le Guen}
\credit{Supervision, Writing - Review \& Editing}
\author[1]{Sibo Cheng}
\credit{Supervision, Writing - Review \& Editing}
\author[1]{Marc Bocquet}
\credit{Supervision, Writing - Review \& Editing}
\author[1]{Alban Farchi}
\credit{Supervision, Writing - Review \& Editing}
\author[4]{Patrick Armand}
\credit{Supervision, Writing - Review \& Editing}
\author[1]{Patrick Massin}
\credit{Supervision, Writing - Review \& Editing}

\cortext[cor1]{Corresponding author}

\address[1] {CEREA, ENPC, EDF R\&D, Institut Polytechnique de Paris, \^Ile-de-France, France}
\address[2] {SINCLAIR AI Laboratory, Saclay, \^Ile-de-France, France}
\address[3] {EDF R\&D, \^Ile-de-France, France}
\address[4] {CEA, DAM, DIF, F-91297 Arpajon, France}

\begin{abstract}
Air flow modeling at a local scale is essential for applications such as pollutant dispersion modeling or wind farm modeling. To circumvent costly Computational Fluid Dynamics (CFD) computations, deep learning surrogate models have recently emerged as promising alternatives. However, in the context of urban air flow, deep learning models struggle to adapt to the high variations of the urban geometry and to large mesh sizes.
To tackle these challenges, we introduce Anchored Branched Steady-state WInd Flow Transformer (AB-SWIFT), a transformer-based model with an internal branched structure uniquely designed for atmospheric flow modeling.
We train our model on a specially designed database of atmospheric simulations around randomised urban geometries and with a mixture of unstable, neutral, and stable atmospheric stratifications. 
Our model reaches the best accuracy on all predicted fields compared to state-of-the-art transformers and graph-based models. Our code and data is available at https://github.com/cerea-daml/abswift.
\end{abstract}





\maketitle

\section{Introduction}

Microscale modeling of the atmospheric flow is important for several applications. However, some applications may require a large number of simulations, such as design optimization of wind farm performances \citep{Ivanell2025}, long-term plant growth under solar panels \citep{Joseph2025}, or real-time modeling and solving inverse problems, such as pollutant dispersion \citep{Tognet2015}. While Computational Fluid Dynamics (CFD) can be used to model the atmospheric flow over complex urban areas \citep{Saturne2025}, CFD simulations can be slow and expensive when high mesh refinement is necessary. This makes them cumbersome when many simulations or fast responses are needed.

Machine learning has recently gained attraction as a viable option to develop fast surrogates \citep{Karniadakis2021}. However, local air flows present several complex challenges for machine learning surrogates. Firstly, for urban areas, the geometry may represent any building shapes and layouts, which can vary significantly. Secondly, deep learning approaches tend to struggle to scale to large meshes required by real-case scenarios. Finally, flow behavior in the atmospheric boundary layer is influenced by atmospheric stratification stability, which modifies the turbulence level in the flow and must be taken into account \citep{Hanna1982}.

To tackle these challenges, we propose Anchored-Branched Steady-state WInd Flow Transformer (AB-SWIFT), a model designed for atmospheric flow modeling. Our model includes an internal branched structure adapted to the multiple components of microscale atmospheric simulations. This allows to flexibly and expressively take into account terrain topology as well as complex obstacles affecting the flow. Our model also takes as inputs vertical meteorological profiles, an important driver of the simulations. This option enables various flow conditions, such as atmospheric stratification stability, with a higher number of degrees of freedom, without being restrained to a specific parameterization of the meteorology.

Our main contributions are as follows:

\begin{itemize}
    \item We present AB-SWIFT, the first transformer based neural operator dedicated to atmospheric flows at a local scale.
    \item We train our model with a new dataset of atmospheric flows in urban areas with different building layouts on flat terrain. Our dataset is also the first to include various atmospheric stratification stabilities.
    \item Our model achieves the best accuracy relative to state-of-the-art Transformers and Graph Neural Networks baselines.
    \item Our code and data are available at https://github.com/cerea-daml/abswift.
\end{itemize}

\subsection{Related works}

\paragraph{Neural operators}

Deep learning models have recently gained traction as surrogate models of physical systems. In particular, data-driven neural surrogates attempt to learn from data the solution operator $\hat S$ that maps inputs $A$ to outputs $U$ \citep{Cheng2025}. For instance, in a time-independent setting, $A$ can represent simulation parameters and geometry, and $U$ can represent 3D stationary volumic fields or integrated scalar quantities. Neural operators \citep{Kovachki2023} attempt to learn directly in the function space, where $A$ and $U$ represent continuous functions. In practice, only access discretized representations of $A$ and $U$ are available. A key requirement to define a neural operator is then the independence to the discretization scheme of $A$ and $U$ and convergence with the discretization resolution.

Most surrogate models, including ours, follow the \textit{encode-process-decode} scheme. This approach decomposes $\hat S$ into an encoder $\hat E$, a processor $\hat P$, and a decoder $\hat D$ so that: $\hat S = \hat D \circ \hat P \circ \hat E$. The encoder transforms physical inputs into \textit{latent tokens}, i.e. scalar-valued vectors of a fixed dimension. The processor can then operate in the dimensional space of the latent tokens, called the \textit{latent space}, and is usually designed to carry the majority of the computational burden. Finally, the decoder maps tokens back into the physical space to predict the desired quantities of interest.

\paragraph{Deep learning surrogates for urban atmospheric flow}

Several past contributions focus on modeling 3D atmospheric flow in built-up environments. \citet{Villeroche2026} use a Multi-Layer Perceptron (MLP) combined with a simple physical model to emulate the air flow around a simple industrial power plant for different wind directions. However, this approach cannot generalize to new geometries. 

\citet{Kastner2023} use a Generative Adversarial Network (GAN) to predict several air flow variables around different obstacle geometries, up to a complex ensemble of buildings representing a lifelike neighborhood. However, this approach requires interpolating unstructured CFD results into a regular 3D grid, and mapping physical fields to colormaps, which results in information and accuracy losses. 

\citet{Shao2023} use Graph Neural Networks (GNN) \citep{Pfaff2020, Brandstetter2022} to model the air flow around various urban topologies for small ensembles of randomly placed buildings. \citet{Liu2023} refine this work by introducing multiple scales in their graph and by partitioning the computation over several sub-graphs, allowing scalability to meshes of about 2 million points. \citet{Shao2024} further improve their model and couple the wind velocity prediction with the prediction of the dispersion of a passive pollutant. However, this GNN-based approach has several limitations.
Firstly, to predict the steady-state, they use a time-iterative approach that starts from an initial state and iterates until it reaches the desired steady state. This results in error accumulation over time steps during inference, and in extra computational time as the model must be trained on all intermediate time steps, and must also be iterated over several steps during inference.
Secondly, GNN are memory-heavy, and do not scale well with the number of mesh points. While partitioning the graph allows for improved scalability, it also increases the computational cost and does not scale well to tens or hundreds of millions of points, which is required by challenging CFD computations. Finally, as the used graphs are built from the mesh connectivity, their models are dependent on the mesh resolution and are not neural operators.

\paragraph{Transformer neural operators for steady-state CFD simulations}

Recently, transformer models have been successfully used as neural operators. Transformers take as inputs a sequence of data of variable length, which can easily represent unstructured simulation data as point clouds \citep{Cheng2025}. The attention mechanism of transformers weighs the "importance" of each element of the input sequence with respect to other elements to produce a new output. As all points of the sequence interact with each other, including elements far from each other in the physical coordinates space, transformers tend to be much better at modeling long distance interactions than other more local approaches such as GNNs and Convolutional Neural Networks. Furthermore, \citet{Kovachki2023, Cao2021} showed that attention can be viewed as a Monte Carlo approximation of a learnable integral operator on the simulation domain. \citet{Calvello2024} demonstrated the universal approximation theorem for transformer-based neural operator, i.e. that they can learn to represent any operator acting between input and output functions defined over the compact simulation domain.

However, computing attention on all mesh points requires quadratic complexity, which is hard to scale to extremely fine meshes of millions to hundreds of millions of points. Transolvers \citep{Wu2024, Luo2025} overcome this limitation by computing a reduced number of physical modes, and only computing attention between them, resulting in a complexity quadratic with the number of modes and not with the number of mesh points. Universal Physics Transformers (UPT) \citep{Alkin2024} and Anchored-Branched Universal Physics Transformers (AB-UPT) \citep{Alkin2025} drastically reduce the complexity by massively under-sampling mesh points and by leveraging anchor attention which restricts the attention mechanism to key points. Additionally, they highlight the potential for transformers to be completely mesh-free at inference time, since model inputs are point clouds, which can be obtained without building a mesh. Finally, they also introduce a "branched" approach, where different parts, or branches, of the model are dedicated to the different inputs and outputs of the problem, such as the geometry, the surfacic and the volumic data.

We provide an overview of the different attention mechanisms used in this work in Appendix \ref{secA:ml}.

\subsection{Atmospheric stability}

In the troposphere (lowest $\qty{10}{\km}$ above ground), pressure and temperature vary with altitude. Depending on the rate of these variations, a particle of air subject to a vertical displacement from its initial position can either go back to its original position, stay in its new position or move further up from its original position. Those three cases correspond respectively to a \textit{stable, neutral}, or \textit{unstable} atmospheric stratification. Stable stratifications are characterized by weaker vertical mixing and turbulence, and longer wakes behind obstacles, while unstable stratifications have stronger turbulence by orders of magnitudes, strong vertical mixing, and relatively short wakes. These characteristics strongly impact how a pollutant plume is dispersed. For more detailed information on these phenomena, readers are referred to \citet{Hanna1982}.

The atmospheric pressure and temperature gradients, and thus the overall stability, depend on the local meteorological conditions. In this study, we use $1 /\Lmo$, the inverse of Monin-Obukhov length \citep{Monin1954} to parameterize the atmospheric stability and to compute meteorological profiles. A negative value of $ 1 /\Lmo$ corresponds to an unstable stratification, while a positive value corresponds to a stable one. Finally, values near zero corresponds to a neutral stratification. We also vary the ground roughness length $z_0$, which quantifies the mechanical turbulence created by the interaction of the air flow with ground textures or unmodeled small patterns. 
While a higher roughness makes a stable stratification more stable, and an unstable stratification more unstable, the impact is relatively small in the range of values considered in this study (i.e. $z_0 < \qty{1}{\m}$) \citep{Hanna1982}. Furthermore, the roughness value does not change the type of the stratification (i.e. stable, neutral or unstable).

Furthermore, we also use the potential temperature $\theta$, defined as the temperature that a particle of air would reach if adiabatically moved to a reference pressure. Unlike the real temperature in the absence of local thermal inversion, $\theta$ decreases with the altitude under unstable stratification conditions, increases under stable conditions, and is constant in neutral conditions.

\section{AB-SWIFT model for atmospheric flow prediction}
\label{sec:model-description}

We introduce AB-SWIFT, a model dedicated to atmospheric flow prediction. To motivate the model design, we first specify the desired criteria (Section \ref{sec:arch-desiderata}), and then describe the building blocks of the architecture (Section \ref{sec:arch-components}). Finally we review the main model hyperparameters (Section \ref{sec:model-hyperparams}).

\subsection{Architecture desiderata}
\label{sec:arch-desiderata}

3D atmospheric flows depend on several inputs: obstacles shapes and positions, terrain topology, and meteorological conditions, i.e. physical context. While obstacles and topology are 3D shapes localized in space, meteorological conditions constitute a global input that impacts the entire behavior of the flow.
Our desired architecture should be able to account for these various inputs. Additionally, a requirement for our architecture would be its ability to predict 3D volumetric information defined on an irregular grid, as the model will be trained on unstructured CFD data.

This leads to the following architectural criteria:
\begin{itemize}
    \item The model should be able to generalize across multiple geometries, each defined over different unstructured meshes.
    \item The architecture must reflect various inputs and how they impact - either locally or globally - the flow. 
    \item Encoder and decoder should naturally be different, since we encode surfacic data (either obstacles or terrain) while we decode volumic data.
    \item The model should reflect the physical dependency of the flow to geometry and volume interactions.
    \item The processor and decoder should be scalable to a very high number of points, as state-of-the-art CFD models often have volumetric meshes containing millions to hundreds of millions of points.
\end{itemize}

\begin{figure}[h]
    \centering
    \includegraphics[width=1.0\linewidth]{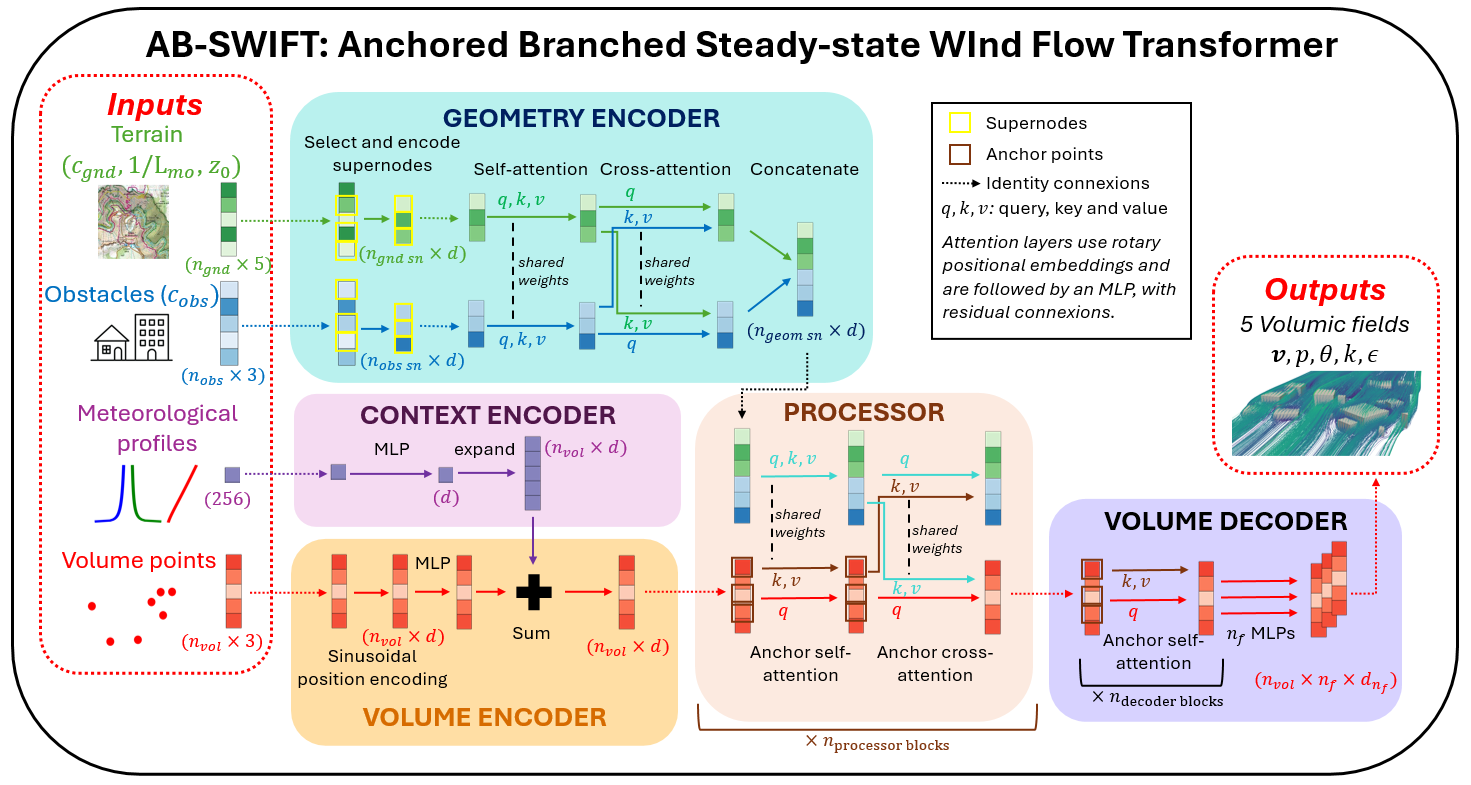}
    \caption{Our proposed model architecture. AB-SWIFT separately encodes the terrain and obstacles to yield one embedded geometry sequence, and encodes the physical context and the volume prediction points. The geometry and encoded prediction points are then processed together. Finally the decoder predicts physical fields from processed latent states of volume points.} 
    \label{fig:model_architecture}
\end{figure}

To fulfill these criteria, we propose an architecture called AB-SWIFT, for Anchored Branched Steady-state WInd Flow Transformer, as illustrated in Figure \ref{fig:model_architecture}. 
AB-SWIFT is an adaptation of AB-UPT, a transformer-based neural operator which takes as inputs and outputs unstructured point clouds. This allows to flexibly encode various geometries and work on unstructured CFD meshes.
To account for the different inputs characterizing atmospheric flows, AB-SWIFT adapts AB-UPT's structure through a redesigned encoder and decoder. AB-SWIFT's geometric encoder treats separately the two formats of geometrical inputs, i.e. the terrain and the obstacles, and produces a single compressed latent sequence representing both. An encoding of the meteorological profiles is also introduced to better account for the meteorological conditions. Additionally, as the atmospheric flow to predict consists of several variables (spatial fields) with different statistical distributions, the final decoder stage is split for each predicted field. The different components are explained in the following subsection.

\subsection{Detailed model components}
\label{sec:arch-components}
\paragraph{Geometry encoder}

The first step is to encode the geometry of the problem, which consists of a cloud of $n_\mathrm{obs}$ points defining the obstacles, and a cloud of $n_\mathrm{gnd}$ points defining the terrain. Following the approach of \citet{Alkin2025}, we assume that a coordinate-based description is sufficient to encode geometric information, and do not include additional geometric features such as surface normals in either point cloud. However, in the terrain point cloud, we attach parameters $1/\Lmo$ and $z_0$ as additional features, as, in the CFD simulations, they directly impact boundary conditions between the ground and the flow, and hence the impact of the terrain on the latter, which should here be reflected. This design choice will also allow to model terrains with inhomogeneous roughness in future studies.

In order to encode information into reduced sequences of latent tokens, we use supernodes embedding layers \citep{Alkin2024} on both point clouds. These layers correspond to randomly selecting a predefined number of \textit{supernodes} from the point cloud. Each supernode then encodes local information from all neighbors present within a radius $r$ using a message-passing layer, i.e. a mean of MLP outputs on neighboring points within the given radius.

We apply separate supernode encoding layers to the terrain and obstacles point clouds, with radii $r_\mathrm{obs}$ for the obstacles and $r_\mathrm{gnd}$ for the terrain. This yields latent sequences of shapes $(n_\mathrm{obs\text{-}sn}, d)$ and $(n_\mathrm{gnd\text{-}sn}, d)$, with $n_\mathrm{obs\text{-}sn}$ and $n_\mathrm{gnd\text{-}sn}$ the chosen numbers of supernodes, and $d$ the hidden dimension of the model's latent states. We further process each layer with a self-attention transformer block, with shared weights. A cross-attention layer is then used to allow sequences to interact with each other. Finally, terrain and obstacle sequences are concatenated into a single geometry sequence representing the full geometry of the problem. With $n_\mathrm{geom\text{-}sn} = n_\mathrm{gnd\text{-}sn} + n_\mathrm{obs\text{-}sn}$, the obtained geometry sequence has shape $(n_\mathrm{geom\text{-}sn}, d)$.

With $\concat$ denoting concatenation, the pseudocode of the forward pass of the geometry encoder is given in Algorithm \ref{alg:geom-enc}.

\begin{algorithm}[h]
    \begin{algorithmic}
      \Input
      \State $c_\mathrm{obs}$   Coordinates of the obstacles point cloud ($\in \R^{n_\mathrm{obs},3}$)
      \State $c_\mathrm{gnd}$   Coordinates of the terrain point cloud ($\in \R^{n_\mathrm{gnd},3}$)
      \State $1 / \Lmo, z_0$   Inverse Monin-Obukhov length and ground rugosity ($\in \R^{n_\mathrm{gnd}}$)
      \EndInput
      \Code
        \State $\ugnd \gets \mathrm{supernode\_encoding_\mathrm{gnd}}(c_\mathrm{gnd} \concat 1 / \Lmo \concat z_0)$
        \State $\uobs \gets \mathrm{supernode\_encoding_\mathrm{obs}}(c_\mathrm{obs})$
        \State $\ugnd \gets \mathrm{self\text{-}attention}(\ugnd)$
        \State $\uobs \gets \mathrm{self\text{-}attention}(\uobs)$
        \State $\ugnd \gets \mathrm{cross\text{-}attention}(\ugnd, \uobs)$
        \State $\uobs \gets \mathrm{cross\text{-}attention}(\uobs, \ugnd)$
        \State $\ugeom \gets \ugnd \concat \uobs $
      \EndCode
      \Output
      \State $\ugeom$   Embedded geometry sequence ($\in \R^{n_\mathrm{geom\text{-}sn},d}$)
      \EndOutput
    \end{algorithmic}
    \caption{Pseudocode of the geometry encoder.}
    \label{alg:geom-enc}
\end{algorithm}

\paragraph{Context encoding}

While atmospheric stability has been parameterized in this study by two scalars $\Lmo$ and $z_0$ with a specific choice of meteorological profile functions (see Appendix~\ref{secA:dataset}), this choice is not universal. In general, different sets of parameters can be chosen, and meteorological profiles could be either derived from other universal functions \citep[such as][]{Carl1973, Hartogensis2005}, pre-computed with 1D CFD simulations \citep{Ivanell2025}, or even measured experimentally \citep{Ferrand2025}.

To account for all possible meteorological conditions, our model encodes meteorological profile functions instead of scalar values of meteorological parameters. We use vertical profiles of the velocity $v$, of the turbulent kinetic energy $k$ and the turbulent kinetic energy dissipation rate $\epsilon$ and of the potential temperature $\theta$, discretised over $64$ vertical levels and flattened into a single array. Then, we embed this information into a latent token space of dimension $d$ using an MLP with a single hidden layer of size $d$ and a GeLU activation function:
\begin{equation}
    \label{eq:profile-encoder}
    u_\mathrm{context} = \mathrm{MLP}\left(v_\mathrm{profile} \concat \theta_\mathrm{profile} \concat k_\mathrm{profile} \concat \epsilon_\mathrm{profile} \right).
\end{equation}

In this study, we precompute the profiles from $\Lmo$ and $z_0$ using the same universal functions as those used in the CFD setup of the data generation. This corresponds to \citet{Hoegstroem1988}' universal functions in unstable stratifications, \citet{Chenge2005}' functions in stable stratifications, and logarithmic profiles in neutral stratifications. We note that computing the profiles is a very fast 1D calculation which cost is negligible compared to the overall computational cost of the model.

\paragraph{Volume encoding}

The $n_\mathrm{vol}$ volume prediction points are encoded using sinusoidal embeddings \citep{Vaswani2017}, followed by an MLP. In order to incorporate physical context information into each volume point, the latent state of the context token is repeated $n_\mathrm{vol}$ times and summed with the latent state of the volume point sequence. The corresponding pseudocode is given in Algorithm \ref{alg:vol-enc}.

\begin{algorithm}[h]
    \begin{algorithmic}
      \Input
      \State $c_\mathrm{vol}$   Coordinates of the volume point cloud ($\in \R^{n_\mathrm{vol},3}$)
      \State $u_\mathrm{context}$   Embedded context information ($\in \R^d$)
      \EndInput
      \Code
        \State $\uvol \gets \mathrm{sin\_embedings}(c_\mathrm{vol})$
        \State $\uvol \gets \mathrm{MLP}(\uvol)$
        \State $u_\mathrm{context} \gets \mathrm{expand\_copy}(u_\mathrm{context}, n_\mathrm{vol})$
        \State $\uvol \gets \uvol + u_\mathrm{context}$
      \EndCode
      \Output
      \State $\uvol$   Embedded volume coordinates ($\in \R^{n_\mathrm{vol},d}$)
      \EndOutput
    \end{algorithmic}
    \caption{Pseudocode of the volume points encoding.}
    \label{alg:vol-enc}
\end{algorithm}

\paragraph{Processor}

Our processor consists of $n_\mathrm{processor\text{-}blocks}$ distinct physics blocks \citep{Alkin2025}, which further process the geometry and volume sequences. Each physics block constitutes a self-attention transformer applied to both sequences, followed by a cross-attention Transformer to allow information to flow between both sequences. To ensure scalability to a high number of volumic points, anchor attention is used in the volume branch for both the self and cross-attention Transformers, with $n_{\mathrm{vol\text{-}anchor}}$ anchor points randomly selected among the $n_\mathrm{vol}$ points with uniform probability distribution. The corresponding pseudocode is given in Algorithm \ref{alg:processor}.

\begin{algorithm}[h]
    \begin{algorithmic}
      \Input
      \State $\ugeom$   Embedded geometry sequence ($\in \R^{n_\mathrm{geom\text{-}sn},d}$)
      \State $\uvol$   Embedded volume sequence ($\in \R^{n_\mathrm{vol},d}$)
      \State $idx_\mathrm{vol\text{-}anchor}$   Indexes of the volume anchor points ($\in \R^{n_\mathrm{vol\text{-}anchor}}$)
      \EndInput
      \Code
        \For{$i$ in range($n_\mathrm{processor\text{-}blocks})$:}
            \State $\ugeom \gets \mathrm{anchor\text{-}self\text{-}attention_i}(\ugeom, \ugeom)$
            \State $\uvol \gets \mathrm{anchor\text{-}self\text{-}attention_i}(\uvol, \uvol [idx_\mathrm{vol\text{-}anchor}])$
            \State $\ugeom \gets \mathrm{anchor\text{-}cross\text{-}attention_i}(\ugeom, \uvol [idx_\mathrm{vol\text{-}anchor}])$
            \State $\uvol \gets \mathrm{anchor\text{-}cross\text{-}attention_i}(\uvol, \ugeom)$
        \EndFor
      \EndCode
      \Output
      \State $\uvol$   processed volume sequence ($\in \R^{n_\mathrm{vol},d}$)
      \EndOutput
    \end{algorithmic}
    \caption{Pseudocode of the processor.}
    \label{alg:processor}
\end{algorithm}

\paragraph{Decoder}

Obtained latent states on volume points are processed independently from the geometry branch with $n_\mathrm{decoder\text{-}blocks}$ self-attention transformers, with anchor attention used for scalability. Finally, independent MLPs are used to decode each physical field from latent states of volume points. Each MLP has a single hidden layer of hidden size $4d$ and GeLU activation functions. The corresponding pseudocode is given in Algorithm \ref{alg:decoder}.

\begin{algorithm}[h]
    \begin{algorithmic}
      \Input
      \State $\uvol$   Processed volume sequence ($\in \R^{n_\mathrm{vol},d}$)
      \State $idx_\mathrm{vol\text{-}anchor}$   Indexes of the volume anchor points ($\in \R^{n_\mathrm{vol\text{-}anchor}}$)
      \EndInput
    \Code
        \For{$i$ in range($n_\mathrm{decoder\text{-}blocks})$:}
            \State $\uvol \gets \mathrm{anchor\text{-}self\text{-}attention_i}(\uvol, \uvol [idx_\mathrm{vol\text{-}anchor}])$
        \EndFor
        \State $\textbf{v}_\mathrm{pred} \gets \mathrm{MLP}_\mathrm{vel}(\uvol)$
        \State $p_\mathrm{pred}          \gets \mathrm{MLP}_\mathrm{p}(\uvol)$
        \State $\theta_\mathrm{pred}     \gets \mathrm{MLP}_\mathrm{\theta}(\uvol)$
        \State $\log k_\mathrm{pred}     \gets \mathrm{MLP}_\mathrm{k}(\uvol)$
        \State $\log \epsilon_\mathrm{pred} \gets \mathrm{MLP}_\mathrm{\epsilon}(\uvol)$
    \EndCode
    \Output
      \State $\textbf{v}_\mathrm{pred}, p_\mathrm{pred}, \theta_\mathrm{pred}, \log k_\mathrm{pred}, \log \epsilon_\mathrm{pred}$  Predicted fields ($\in \R^{7,n_\mathrm{vol}}$)
    \EndOutput
    \end{algorithmic}
    \caption{Pseudocode of the Decoder.}
    \label{alg:decoder}
\end{algorithm}

\subsection{Model hyperparameters}
\label{sec:model-hyperparams}

Table \ref{tab:model-hyperparams} summarizes the main model hyperparameters, as well as hyperparameters characterizing the transformer blocks used. AB-SWIFT specific parameters were tuned though trial and errors, starting with parameter values similar to what is proposed by \citet{Alkin2025} for the Drivaernet++ dataset. A finer tuning of these hyperparameters can be done, but would require significant computational resources, and is left to future work.

Hyperparameters defining transformer blocks were not tuned and correspond to parameters of a ViT-tiny \citep{Dosovitskiy2020}. Following standard design choices, each transformer block is made up of a multi-head attention layer followed by an MLP, with residual connection between each sublayers. MLPs have a single hidden layer of size $4d$ and GeLU activation functions. Additionally, axis rotary positional embeddings (RoPE) \citep{Su2024} are used to embed positional information in the attention mechanism.

\begin{table}[h]
    \centering
    \begin{tabular}{llr} \toprule
         Name & Description & Value \\ \midrule
         \multicolumn{3}{c}{AB-SWIFT specific parameters} \\ \addlinespace
        $d$ & Hidden dimension of the latent states & 192 \\
        $n_\mathrm{obs}$ & Number of points describing obstacles & 4096 \\
        $n_\mathrm{gnd}$ & Number of points describing the terrain & 4096 \\
        $n_\mathrm{obs\text{-}sn}$ & Number of obstacles supernodes & 1024 \\
        $r_\mathrm{obs}$ & Radius for obstacles' supernodes pooling & 1 \\
        $n_\mathrm{gnd\text{-}sn}$ & Number of terrain supernodes   & 1024 \\
        $r_\mathrm{gnd\text{-}sn}$ & Radius for terrain supernodes pooling & 5 \\
        $n_\mathrm{vol\text{-}anchor}$ & Number of volume anchor point & 8192 \\
        $n_\mathrm{processor\text{-}blocks}$ & Number of processor blocks & 3 \\
        $n_\mathrm{decoder\text{-}blocks}$ & Number of self-attention transformer blocks in the decoder & 4 \\
        \midrule 
        \multicolumn{3}{c}{Transformer layers parameters} \\ \addlinespace
         & Number of attention heads & $3$ \\
         & Positional embeddings & RoPE \citep{Su2024} \\
         & Number of hidden layer in the MLP & 1 \\
         & Hidden size of the MLP & 4d \\
         & Activation function of the MLP & GeLU \\
        \bottomrule
    \end{tabular}
    \caption{AB-SWIFT hyperparameters and used values.}
    \label{tab:model-hyperparams}
\end{table}

\section{Results and comparison to other models}

\subsection{Dataset}
\label{sec:database_generation}

In order to train and evaluate models, we propose a new database of time-averaged steady-state atmospheric flows around various obstacle geometries and for different atmospheric stability conditions. CFD simulations were carried out using code\_saturne \citep{Saturne2025}. We present here a brief overview of the dataset. Additional information on the used generation method and on the CFD setup can be found in Appendix~\ref{secA:dataset}.

\begin{figure}[h]
    \centering
    \includegraphics[width=0.9\linewidth]{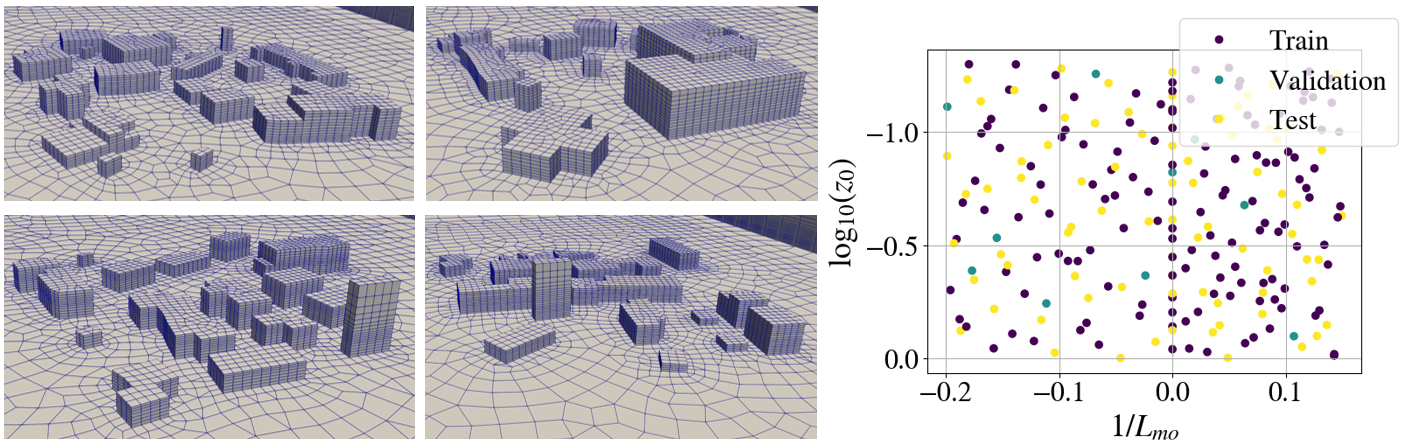}
    \caption{Left: $4$ building configuration, taken from the $210$ different configurations present in our dataset. Right: used values of $1/\Lmo$ and $z0$. Purple points represent the training split, blue the validation split, and yellow the test split.}
    \label{fig:dataset}
\end{figure}

Each sample has a square built area were buildings are located, upstream of a long downwake area allowing the flow to settle. Obstacle geometries are determined by randomly sampling multiple buildings from an ensemble of predefined building shapes, and placing them in the built area without overlap to obtain various urban-like geometries. The wind orientation is constant and comes from the west side of the simulation. Additionally, each sample has a different atmospheric stratification stability, parameterized by $1/ \Lmo$ and $z_0$. Figure \ref{fig:dataset} shows $4$ example geometries and the sampled values of $1/\Lmo$ and $z_0$. Finally, for each sample, output variables are steady-state atmospheric flow quantities defined at each mesh cell center: the velocity field $\mathbf{v}$, the potential temperature $\theta$, the pressure variation $p$, the turbulent kinetic energy $k$ and the turbulent kinetic energy dissipation rate $\epsilon$. Depending on the sampled geometry, samples have between $\num{50000}$ to $\num{200000}$ cells.

The dataset consists of $\num{228}$ simulations, split as $138$ training samples, $10$ validation samples, and $70$ testing samples. All atmospheric stability categories are represented in each subset. Additionally, $3$ geometries were repeated, for $6$ different stabilities per repeated geometry, to avoid having a unique pairing between geometries and stabilities in the training dataset. A precise breakdown of the stabilities for each split is shown in Table \ref{tab:dataset-split}. In all subsets, the neutral class is less sampled because it corresponds to a narrower range of values of $1 / \Lmo$, while the associated probability distribution is uniform.

\begin{table}[h]
    \centering
\begin{tabular}{ccccc} \toprule
        & unstable
        & neutral
        & stable
        & total \\ \cmidrule(lr){2-5}
    training       & 53 & 15 & 70 & 138 \\
    validation  & 6  & 1  & 3  & 10  \\
    test        & 35 & 11 & 34 & 80  \\ \bottomrule
\end{tabular}
    \caption{Dataset split between training, validation and test samples with stability repartition for each split.}
    \label{tab:dataset-split}
\end{table}

\subsection{Evaluation metrics}

In order to evaluate the surrogate model predictions, three metrics are computed over each predicted physical field. In this study, we use the normalized mean square error (NMSE), the L1 error, and the L2 error. The NMSE is normalized by the variance of a given field for a simulation, and L1 and L2 errors are normalized by the L1 and L2 norms of the field respectively.  
Hence, NMSE tends to be more relevant when measuring fields that have a small standard deviation compared to their absolute value, such as the potential temperature, for which the normalizations of the L1 and L2 metrics would lead to artificially small error values. By contrast, NMSE will appear artificially small when the standard deviation of a field is high compared to its absolute value. Consequently, these metrics are complementary to properly evaluate the prediction's accuracy.

However, as the potential temperature is spatially homogeneous in neutral stratifications, NMSE for $\theta$ is only computed over unstable and stable stratifications to avoid dividing by $0$ when normalizing with the variance.

For a given simulation sample $i$, with $y^\mathrm{true}_i$ the CFD ground truth of field $y_i$ and $y^\mathrm{pred}_i$ the surrogate model prediction, the metrics for a single simulation are defined as:
\begin{subequations}
    \label{eq:metrics}
    \begin{align}
        & \textrm{NMSE}_i = \frac{\mathrm{mean}\left(\left(y^\mathrm{true}_i - y^\mathrm{pred}_i\right)^2\right)}{\mathrm{var}(y^\mathrm{true}_i)},
        \\
        & \textrm{L}1\_\mathrm{err}_i = \frac{\left\|y^\mathrm{true}_i - y^\mathrm{pred}_i\right\|_1}{\left\|y^\mathrm{true}_i\right\|_1},
        \\
        & \textrm{L}2\_\mathrm{err}_i = \frac{\left\|y^\mathrm{true}_i - y^\mathrm{pred}_i\right\|_2}{\left\|y^\mathrm{true}_i\right\|_2},
    \end{align}
\end{subequations}
where mean and var operators are the arithmetic mean and variance computed over the simulation domain and field components.

The metrics of a given surrogate model are computed for all simulations in the test set (Table 2), and their average and standard deviation over the whole set are analyzed.

\subsection{Benchmark}

We train AB-SWIFT on our dataset and compare it with recent competitive baselines: AB-UPT \citep{Alkin2025}, GAOT \citep{Wen2025}, Transolver \citep{Wu2024} and Bi-stride multiscale MeshGraphNet (BSMGN) \citep{Cao2023}. Models setup are summarized in Table \ref{tab:benchmark-inputs-outputs}, and are as follows. 

For AB-UPT, the geometry point cloud represents both terrain and obstacles. Additionally simulation parameters $1/\Lmo$ and $z_0$ are added as additional features in the geometry point cloud. 

For non-branched architectures Transolver, GAOT and BSMGN, only the volume point cloud is fed into the model, with simulation parameters as additional features, and with a one-hot encoded feature to distinguish points near the ground, near the different domain boundaries, near the obstacles, or in the volume. In the case of the GNN-based architecture, BSMGN, a graph is generated from the CFD mesh connectivity.

For all models, the outputs are 5 fields characterizing the flow: $\textbf{v}, \theta, p, k, \epsilon$. In order to facilitate the learning process across multiple orders of magnitude, models are trained to predict base 10 logarithm $\log k$ and $\log \epsilon$ instead of $k$ and $\epsilon$ as those variables are positive and vary considerably across different orders of magnitudes depending on atmospheric stability.

Coordinate-based inputs are normalized between $0$ and $\num{1000}$, while scalar features and fields are standardized (zero-mean and unit-variance) across the training dataset. One-hot encoded features are not normalized.

\begin{table}[h]
    \centering
    \label{tab:benchmark-inputs-outputs}
    \begin{tabular}{c m{3cm}m{6cm}} \toprule
         & Input format & Input features \\ \midrule
         AB-SWIFT & Terrain, obstacles and volume point clouds & Atmospheric profiles \newline $1/\Lmo$ and $z_0$ in the terrain point cloud \\ \midrule
         AB-UPT & Geometry and volume point clouds & $1/\Lmo$ and $z_0$ in the geometry point cloud \\ \midrule
         GAOT & Volume point cloud & $1/\Lmo, z_0$, one-hot encoded point type \\ \midrule
         Transolver & Volume point cloud & $1/\Lmo, z_0$, one-hot encoded point type \\ \midrule
         BSMGN & Graph from volume mesh connectivity & $1/\Lmo, z_0$, one-hot encoded node type \\ \bottomrule
    \end{tabular}
    \caption{ Inputs format and used features of our different tested models. For all models outputs are $\textbf{v}, \theta, p, \log k, \text{and} \log \epsilon$.}
\end{table}

Model hyperparameters such as the number of layers and the latent state size were chosen so that all models had approximately $6$M trainable parameters, except BSMGN, which had to be reduced to about $0.4$M parameters so that training would fit in memory on a single $40$Gb A100 GPU. When possible, hyperparameters were chosen to match parameter choices in \citet{Alkin2025} for the Drivaernet++ dataset, which is the one most resembling our dataset among those studied in the different papers. Finally, AB-SWIFT, AB-UPT and GAOT were parameterized to have the same number of anchor tokens or latent tokens. Chosen hyperparameters for each model are detailed in Appendix~\ref{secA:params}.

All models were trained for $500$ epochs and with a batch size of $1$. We used a mean square error loss function, with a OneCycle learning rate decay \citep{Smith2019} and a maximum learning rate of $\num{1e-3}$. Mixed precision was used for all models except GAOT due to instabilities numerical instabilities during training. All trainings and inferences were carried out on a single Nvidia A100 40Gb GPU.

Table \ref{tab:model-efficiency} shows the training and inference ressource requirements for each model. AB-SWIFT and AB-UPT are the fastest models to train per epoch, and have the smallest VRAM request. All models can infer in less that $\qty{1}{\s}$, with GAOT being the fastest. The low VRAM requirement of AB-SWIFT and AB-UPT shows that they can computationally scale to meshes of tens of millions of cells, as required in a real-case scenarios.

\begin{table}[h]
    \centering
    \begin{tabular}{m{4cm}m{2.1cm}m{2.1cm}m{2.1cm}} \toprule
         model \newline (\# of trainable parameters) & training time (one epoch) & training VRAM requirement & inference time (one sample) \\
         \midrule
         AB-SWIFT (6.5M)   & $\qty{8}{\s}$   & $\qty{1.21}{Gb}$ & \qty{0.39}{s} \\ 
         AB-UPT (6.5M)     & $\qty{7.5}{\s}$ & $\qty{1.20}{Gb}$ & \qty{0.43}{s} \\ 
         GAOT (6.6M)       & $\qty{17}{\s}$  & $\qty{10}{Gb}$   & \qty{0.23}{s} \\ 
         Transolver (6.0M) & $\qty{27}{\s}$  & $\qty{23}{Gb}$   & \qty{0.31}{s} \\ 
         BSMGN (0.4M)      & $\qty{90}{\s}$  & $\qty{36}{Gb}$   & \qty{0.60}{s} \\
         \bottomrule
    \end{tabular}
    \caption{Resource requirements of all models for training and inference. Training and inference speeds are measured on a A100 GPU with 40Gb of VRAM. Memory cost is shown for a batch size of $1$.}
    \label{tab:model-efficiency}
\end{table}

Table \ref{tab:model-accuracy} shows the obtained metrics computed on the test set. AB-SWIFT consistently has the best accuracy for all considered fields and metrics. AB-UPT and BSMGN have the best accuracy among all literature models, depending on the field and metrics considered.

\begin{table}[h]
    \centering
    \begin{tabular}{cc *{5}{c}} \toprule
        \multicolumn{2}{c}{} & AB-SWIFT (ours) & AB-UPT & GAOT & Transolver & BSMGN \\
        \midrule
         \multirow{3}{*}{$\textbf{v}$} & NMSE & $\textbf{0.3}\pm\textbf{0.4}$ & $0.8\pm1.3$ & $13.1\pm7.4$ & $2.1\pm1.5$ & $\textit{0.8}\pm\textit{1.0}$ \\
         & L1 err & $\textbf{2.2}\pm\textbf{1.3}$ & $5.3\pm3.9$ & $20.4\pm8.0$ & $5.6\pm2.4$ & $\textit{4.6}\pm\textit{2.7}$ \\
         & L2 err & $\textbf{3.7}\pm\textbf{1.8}$ & $\textit{6.2}\pm\textit{3.8}$ & $29.3\pm9.1$ & $9.8\pm3.7$ & $6.3\pm3.2$ \\
         \bottomrule
         \multirow{3}{*}{$p$} & NMSE & $\textbf{2.0}\pm\textbf{3.5}$ & $\textit{5.7}\pm\textit{6.7}$ & $96.8\pm412.3$ & $14.5\pm25.3$ & $6.4\pm9.4$ \\
         & L1 err & $\textbf{9.3}\pm\textbf{7.8}$ & $22.1\pm14.2$ & $62.3\pm76.5$ & $23.9\pm18.6$ & $\textit{20.2}\pm\textit{13.7}$ \\
         & L2 err & $\textbf{9.8}\pm\textbf{9.2}$ & $\textit{18.5}\pm\textit{12.5}$ & $64.2\pm68.4$ & $26.4\pm24.1$ & $18.6\pm14.6$ \\
         \bottomrule
         \multirow{3}{*}{$\theta$} & NMSE* & $\textbf{4.0}\pm\textbf{9.8}$ & $\textit{20.3}\pm\textit{31.6}$ & $3622.9\pm8317.8$ & $26.6\pm37.9$ & $22.0\pm38.1$ \\
         & L1 err & $\textbf{0.1}\pm\textbf{0.1}$ & $0.2\pm0.2$ & $1.5\pm1.3$ & $0.2\pm0.1$ & $\textit{0.2}\pm\textit{0.1}$ \\
         & L2 err & $\textbf{0.1}\pm\textbf{0.1}$ & $0.3\pm0.3$ & $2.3\pm2.0$ & $0.3\pm0.2$ & $\textit{0.2}\pm\textit{0.2}$ \\
         \bottomrule
         \multirow{3}{*}{$\log k$} & NMSE & $\textbf{11.0}\pm\textbf{11.6}$ & $37.0\pm42.4$ & $956.6\pm914.8$ & $35.0\pm20.2$ & $\textit{20.2}\pm\textit{16.1}$ \\
         & L1 err & $\textbf{4.6}\pm\textbf{4.7}$ & $14.9\pm17.6$ & $50.3\pm42.2$ & $9.6\pm9.9$ & $\textit{7.4}\pm\textit{6.8}$ \\
         & L2 err & $\textbf{7.7}\pm\textbf{6.1}$ & $15.8\pm14.0$ & $71.3\pm51.3$ & $15.8\pm11.6$ & $\textit{11.1}\pm\textit{7.8}$ \\
         \bottomrule
         \multirow{3}{*}{$\log \epsilon$} & NMSE & $\textbf{4.3}\pm\textbf{4.9}$ & $19.1\pm27.4$ & $402.3\pm591.1$ & $15.9\pm9.2$ & $\textit{9.0}\pm\textit{9.1}$ \\
         & L1 err & $\textbf{3.4}\pm\textbf{2.3}$ & $18.0\pm26.8$ & $62.9\pm87.8$ & $9.0\pm7.1$ & $\textit{6.7}\pm\textit{5.1}$ \\
         & L2 err & $\textbf{5.7}\pm\textbf{3.0}$ & $17.2\pm21.7$ & $84.0\pm107.1$ & $14.1\pm9.6$ & $\textit{9.4}\pm\textit{5.9}$ \\
         \bottomrule
    \end{tabular}

    \caption{Benchmark results of AB-SWIFT and literature baselines. NMSE is in percent of the variance over a given simulation, and L1 and L2 errors are in percent of the ground truth value. We report mean and standard deviation of metrics computed on the test split of the dataset for all trained models. For all metrics, the \textbf{best} model is shown in bold, and the \textit{second best} in italic. *NMSE for $\theta$ is computed on stable and unstable cases only.}
    \label{tab:model-accuracy}
\end{table}

\begin{figure}[h]
    \centering
    \includegraphics[width=\linewidth]{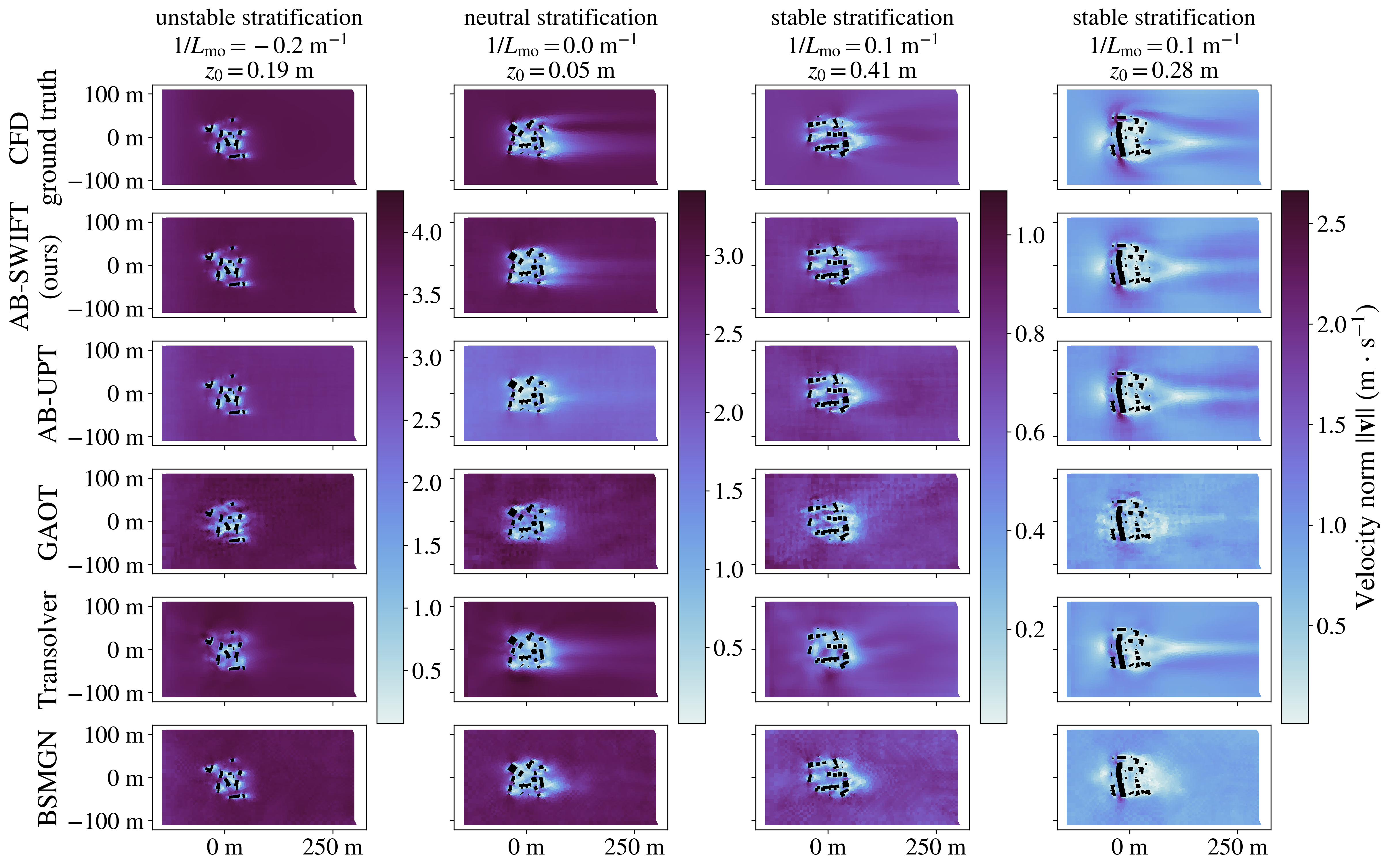}
    \caption{Norm of the velocity field predicted by all models, $\qty{2}{\m}$ above ground, for an unstable stratification (Left), a neutral stratification (Middle left), and two stable stratifications (Middle right and Right). All shown geometries and stability parameters are from the test split of the dataset.}
    \label{fig:models-comp}
\end{figure}

For a more visual comparison, Figure \ref{fig:models-comp} shows the predicted velocities for all models on an unstable, a neutral and two different stable stratifications, with previously unseen geometries. AB-SWIFT, AB-UPT and BSMGN  seem to accurately capture the wake shapes near buildings that are heavily determined by building geometries. By contrast, Transolver struggles in the buildings area, which can be expected as it lacks a dedicated geometry encoding or a local-scale embedding mechanism. Surprisingly, GAOT also struggles at predicting near-field physics, despite its strong emphasis on geometry encoding. 
Finally, while AB-UPT seems good at capturing near-field and long-range wakes, it tends to predict far-field flow incorrectly, which can be interpreted as an incorrect prediction of the stability regime.

This figure also shows that a graph-based BSMGN struggles at predicting long-range wakes. Long distance interactions are often hard to model with GNN based models due to over-smoothing over multiple message-passing \citep{Rusch2023}. By contrast, all transformer-based models except GAOT predict long distance wakes more accurately. 

\begin{figure}[h]
    \centering
    \includegraphics[width=\linewidth]{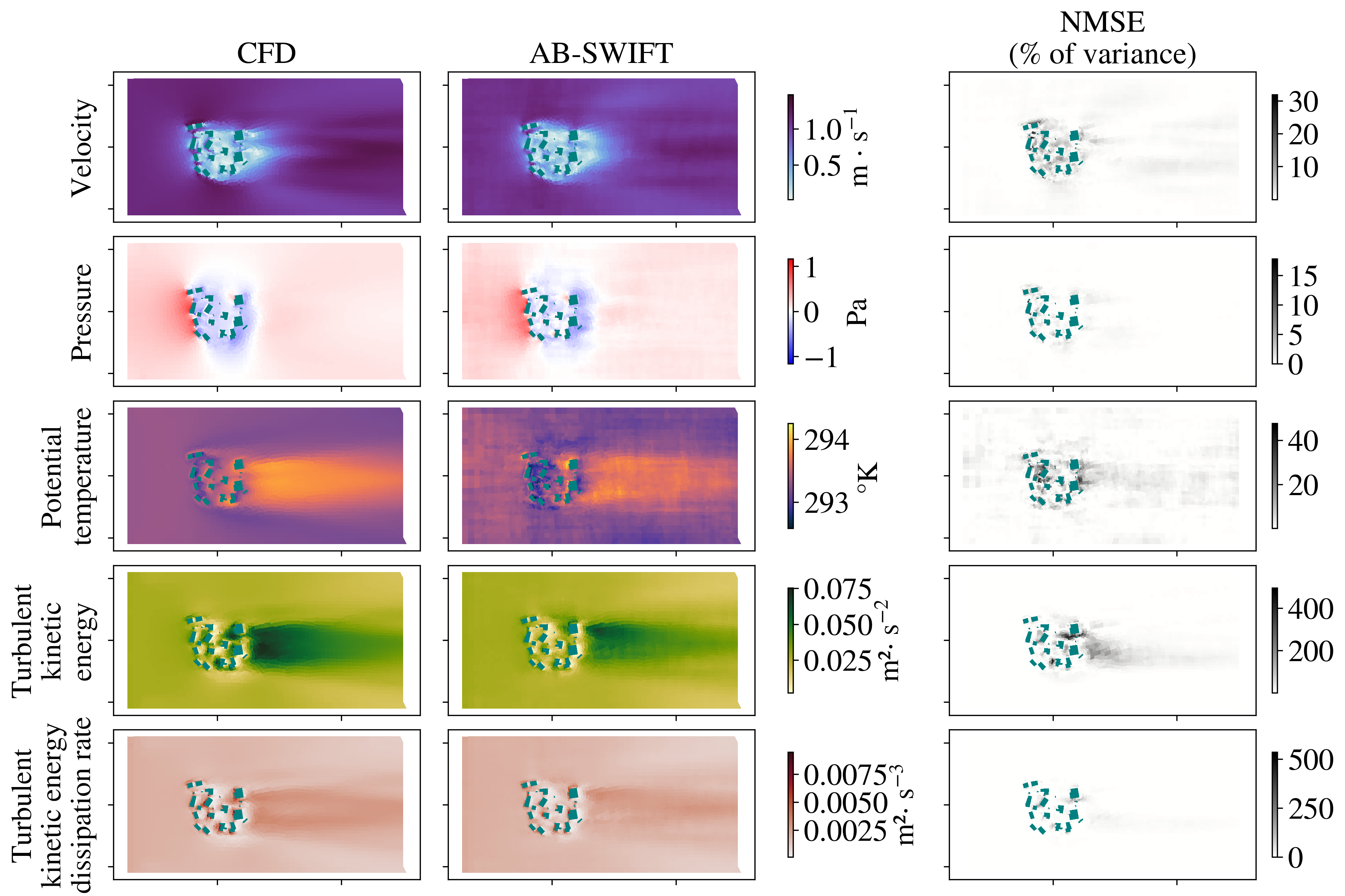}
    \caption{Horizontal slice at height $h=\qty{2}{\m}$ of the fields predicted by AB-SWIFT for a stable stratification ($1/\Lmo=\qty{0.15}{\per\m}, z_0=\qty{0.06}{\m})$. Buildings are shown in teal color.}
    \label{fig:sample-prediction}
\end{figure}

Finally, Figure \ref{fig:sample-prediction} shows the predicted fields with AB-SWIFT for a stable stratification scenario, on a geometry unseen during training. We can observe that all considered fields are correctly predicted, despite their different behaviors. The largest errors are concentrated near the buildings. It can also be pointed out that the potential temperature prediction appears to be very noisy, which we attribute to its low variance for this simulation compared to other simulations present in the dataset.

\section{Ablation study}

In order to evaluate the added value of each step of AB-SWIFT, we run an ablation study. Starting from AB-UPT, we decompose the construction of AB-SWIFT in 4 steps, and present the added value of each step in Table \ref{tab:ablation}. The hyperparameters that are common to all steps remain unchanged.

The first step is merging the geometry and surface branches of AB-UPT. Since we do not aim at predicting surfacic values, we do not need a dedicated surface branch independent of the geometry encoding. Instead, we choose to directly use the geometry sequence obtained from the geometry encoder as latent tokens for the surface branch of the processor.

The second step splits the encoding of the geometry into separate sub-branches for the terrain and for the obstacles. While both terrain and obstacles have the same data structure, they represent different physical objects that should be encoded with separate layers. We replace the single branch self-attention transformer of AB-UPT's geometry encoder with a self-attention and a cross-attention transformer to add cross-branch interactions between terrain and obstacles. Finally, we concatenate both sequences to obtain a single geometry sequence used by the transformer, yielding the geometry encoder presented in Section \ref{sec:model-description}.

The third step is adding the encoding of the meteorological profiles. As profiles impact the entire volume flow, a latent representation of the profiles is fed into all volume points.

The final step concerns the change of the linear decoding layer of AB-UPT into separate, non-linear, MLP for each field. This allows a better separation between the different physical fields we are predicting, allowing for better predictions when fields are not directly correlated, such as pressure and velocity.

\begin{table}[]
    \centering
    \begin{tabular}{cc*{5}{c}} \toprule
        \multicolumn{2}{c}{} & Step 0 & Step 1 & Step 2 & Step 3 & Step 4 \\
        \multicolumn{2}{c}{} & \multicolumn{1}{c}{AB-UPT} &&&& \multicolumn{1}{c}{AB-SWIFT} \\
        \midrule
         \multirow{4}{*}{$\textbf{v}$} & NMSE & $1.5\pm1.5$ & $0.6\pm0.6$ & $0.6\pm1.4$ & $0.4\pm0.5$ & $0.3\pm0.4$ \\ 
         & L1 error & $8.5\pm5.3$ & $3.7\pm2.5$ & $3.5\pm3.5$ & $2.5\pm1.6$ & $2.2\pm1.3$ \\ 
         & L2 error & $8.9\pm4.7$ & $5.1\pm2.6$ & $4.8\pm3.4$ & $4.3\pm2.2$ & $3.7\pm1.8$ \\ 
         \midrule
         \multirow{3}{*}{$p$} & NMSE & $8.5\pm12.6$ & $4.8\pm7.3$ & $3.8\pm6.9$ & $2.7\pm4.5$ & $2.0\pm3.5$ \\ 
         & L1 error & $26.6\pm19.9$ & $16.2\pm13.0$ & $14.1\pm10.7$ & $10.9\pm8.9$ & $9.3\pm7.7$ \\ 
         & L2 error & $21.4\pm16.7$ & $15.5\pm13.1$ & $13.8\pm11.9$ & $11.4\pm10.4$ & $9.8\pm9.2$ \\ 
         \midrule
         \multirow{4}{*}{$\theta$} & NMSE* & $29.0\pm60.8$ & $17.3\pm30.9$ & $12.1\pm20.9$ & $6.5\pm15.2$ & $4.1\pm10.0$ \\ 
         & L1 error & $0.2\pm0.3$ & $0.2\pm0.2$ & $0.2\pm0.2$ & $0.1\pm0.1$ & $0.1\pm0.1$ \\ 
         & L2 error & $0.3\pm0.4$ & $0.2\pm0.2$ & $0.2\pm0.2$ & $0.1\pm0.1$ & $0.1\pm0.1$ \\ 
         \midrule
         \multirow{4}{*}{$\log k$} & NMSE & $80.1\pm119.4$ & $23.5\pm33.6$ & $29.6\pm119.6$ & $11.6\pm10.4$ & $11.0\pm11.3$ \\ 
         & L1 error & $25.5\pm36.8$ & $11.2\pm20.5$ & $9.9\pm14.7$ & $4.3\pm4.4$ & $4.6\pm4.7$ \\ 
         & L2 error & $23.8\pm24.5$ & $12.1\pm12.1$ & $11.3\pm10.2$ & $8.0\pm6.1$ & $7.7\pm6.1$ \\ 
         \midrule
         \multirow{4}{*}{$\log \epsilon$} & NMSE & $39.1\pm59.7$ & $7.6\pm7.1$ & $7.9\pm11.6$ & $5.2\pm5.8$ & $4.3\pm4.9$ \\ 
         & L1 error & $26.5\pm40.0$ & $6.6\pm5.3$ & $7.0\pm6.7$ & $3.6\pm2.5$ & $3.4\pm2.3$ \\ 
         & L2 error & $24.4\pm32.5$ & $8.6\pm5.1$ & $8.5\pm5.8$ & $6.4\pm3.6$ & $5.7\pm3.0$ \\ 
         \bottomrule
    \end{tabular}
    \caption{Ablation study results. Starting from AB-UPT, we build AB-SWIFT step by step. Step 1: we merge the geometry and surface branches. Step 2: we encode separately obstacles and terrain. Step 3: we encode profiles directly. Step 4: we split the final decoder layer per predicted fields, resulting in the final AB-SWIFT proposed model. Reported metrics are the mean and standard deviation for each trained intermediate model over the test split of the dataset. NMSE is in percent of the variance over a given simulation, and L1 and L2 errors are in percent of the ground truth value. }
    \label{tab:ablation}
\end{table}

We observe that all steps lead to a reduction of the error. The final model has errors of less than $10\%$ for all metrics and fields, except for NMSE of $\log k$, that reaches $11\%$. In particular, Step 3, i.e. the addition of the meteorological profiles encoder, significantly reduces the prediction errors on $k$ and $\epsilon$. Considering that those variables represent turbulence, which is strongly related to stratification stability, we can see that these steps help to correct the error on the stability regime initially observed in AB-UPT.

\section{Conclusion}

In this paper, we have presented AB-SWIFT, the first transformer-based neural operator for local scale atmospheric simulations that can model different obstacle geometries and atmospheric stratification stabilities. To the best of the authors' knowledge, this is the first work applying a metamodel for atmospheric flow under variable stratification conditions.

We showed that AB-SWIFT outperforms current state-of-the-art models for the presented atmospheric flow problem, with an error at least twice as small as the second best model for all metrics and fields. Additionally, it has a training efficiency and VRAM footprint comparable to the best state-of-the-art model AB-UPT, suggesting a potential to scale to meshes of up to hundreds of millions of points as demonstrated by \citet{Alkin2025}.

A key limit of AB-SWIFT is its inability to generalize to simulation domains larger than those used during training. This is due to the use of rotary positional encoding, which has been shown to fail to generalize to distance larger than what is seen during training in natural language processing \citep{Press2021}. An interesting perspective would be to incorporate different positional embedding mechanisms, such as ALiBi \citep{Press2021}, which have better generalization properties.

Another appealing perspective would be to build a higher fidelity dataset comprising of much larger and more realistic simulations, as AB-SWIFT has the potential to scale to large meshes.

\appendix

\section{Transformer layers}\label{secA:ml}

This appendix details the different attention mechanisms used in this work, and the structure of the transformer blocks used.

\subsection{Attention mechanism}

The Transformers architecture was first introduced by \citet{Vaswani2017} for natural language processing and has been successfully applied to many areas of machine learning \citep{Dosovitskiy2020, Liu2021}. At its core, Transformers rely on the attention mechanism, which allows us to compute relations between a sequence of latent tokens. In this paper, we focus on three variants of the attention mechanism described in the following.

\textbf{Self-attention:} self-attention corresponds to performing attention within elements of a single sequence. Self-attention computes \textit{all to all} relations between elements of the sequence, and modifies the elements accordingly. 

For a sequence with latent states $u$ made of $n$ elements  of hidden dimension $d$, we define the query $q$, key $k$ and value $v$ tensors, defined as $q = uW_q$, $k=uW_k$, and $v=uW_v$, with $W_q, W_k$ and $W_v$ trainable weights matrices. With the softmax function over matrix of elements $u_{i,j}$ defined as $\mathrm{softmax}(u_{i,j}) = \frac{e^{u_{i,j}}}{\sum_k e^{u_{i,k}}}$, self-attention reads:
\begin{equation}
    \label{eq:self-attn}
    \mathrm{self\text{-}attention}(u) = \mathrm{softmax}\left( \frac{q k^\top}{\sqrt{d}} \right) v. 
\end{equation}

\textbf{Cross-attention:} cross attention corresponds to performing attention between two different sequences. In cross attention, information from a second sequence $u_2$ is fed into the first sequence $u_1$, i.e. sequence 2 \textit{attends to} sequence 1. With $q_1=u_1W_v$, $k_2 = u_2W_k$, and $v_2 = u_2W_v$, Cross attention reads:
\begin{equation}
    \label{eq:cross-attn}
    \mathrm{cross\text{-}attention}(u_1, u_2) = \mathrm{softmax}\left( \frac{q_1k_2^\top}{\sqrt{d}} \right) v_2. 
\end{equation}

\textbf{Anchor attention:} anchor attention is a special attention mechanism introduced by \citet{Alkin2025} for large-scale physical simulations. Anchor attention randomly selects a small subset of the total sequence to act as \textit{anchor points}. Non-anchor points are called \textit{query points}. Anchor points then attend to the entire sequence, including themselves:
\begin{subequations}
    \label{eq:anchor-attn}
    \begin{align}
        & u_\mathrm{anchor} = \mathrm{Sample}(u), \\
        & q = u W_q, k_\mathrm{anchor} = u_\mathrm{anchor} W_k, v_\mathrm{anchor} = u_\mathrm{anchor} W_v, \\
        & \mathrm{anchor\text{-}attention}(u) = \mathrm{softmax}\left( \frac{qk_\mathrm{anchor}^\top}{\sqrt{d}} \right) v_\mathrm{anchor}. 
    \end{align}
\end{subequations}

Compared to self-attention, anchor attention significantly reduces the complexity, from $\mathcal{O}(n^2)$ to $\mathcal{O}(nn_\mathrm{anchor})$, with $n$ the total sequence length. Additionally, query points do not depend on each other, which allows to split the computation into several small batches of length $n_\mathrm{bs}$, reducing memory cost to $\mathcal{O}(n_\mathrm{bs}n_\mathrm{anchor})$. This renders scaling to hundreds of millions of points feasible on a single GPU \citep{Alkin2025}. Finally, this property also allows to infer only on a subpart of the simulation domain if needed, providing further speedups when a computation over the full domain is not required.

\subsection{Rotary position encoding}
In order to provide positional information to the attention mechanism, we need to embed positional information to the query and key tokens. In this work, we used axis rotary positional embeddings (RoPE) \citep{Su2024}. RoPE rotate the query and key features $2$ by $2$ with rotations proportional to their respective positions, such that the attention logits $qk^\top$ are dependent on the relative positions between the query and key tokens. For more information, readers are referred to \citet{Su2024}.

\subsection{Multi-head attention}

Transformers used in machine learning typically uses multiple heads of attention: each multi-head attention layer is the sum of several attention layers, with each sub-layer having its own independent weights:
\begin{equation}
    \mathrm{multi\text{-}head\text{-}attention}(u) = \sum_h \mathrm{attention}_{w_q^h, w_k^h,w_v^h}(u).
\end{equation}

\subsection{Transformer blocks}

We use macro blocks that rely on the different attention mechanisms. Each transformer variant is made up of a multi-head attention layer and an mlp, with residual connexion between each sublayers. Additionally, axis rotary postional embeddings (RoPE) \citep{Su2024} are used in the attention layer.

\section{Additional information on the database and CFD setup}\label{secA:dataset}

This appendix aims to provide detailed information on the database generation, complementarily with Section \ref{sec:database_generation}.

\subsection{Geometry generation and stability parameters sampling}

All the modeled buildings were sampled from shapefiles describing buildings in a urban area in Champs-sur-Marne, \^Ile-de-France, France. Multiple buildings are randomly selected from the dataset, individually or by clusters to preserve realistic arrangements. Selected buildings are iteratively and randomly placed in $\qty{100}{\m} \times \qty{100}{\m}$ square areas, with a condition such that they do not overlap. Each geometry is constituted of around $35$ buildings. We note that more advanced placements can be made to obtain more realistic arrangements, such as the one proposed in the bin-packing algorithm \citep{Shao2024, Zhao2024}. However, this upgrade is left to future work.

In parallel, we sample values of $ 1 / \Lmo$ between $\qty{-0.20}{\per\m}$ and $\qty{0.10}{\per\m}$, covering a wide range of stable, neutral, and unstable stratifications. We also sample values of $z_0$ between $\qty{0.05}{\m}$ and $\qty{1.0}{\m}$, which represents rugosities typical of urban and suburban areas.

\subsection{CFD simulation setup}

\begin{figure}[h]
    \centering
    \includegraphics[width=0.8\linewidth]{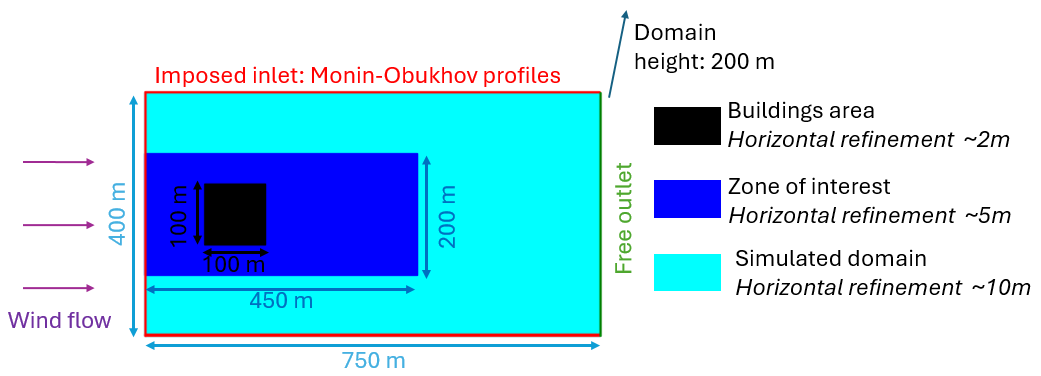}
    \caption{Mesh arrangements used to generate database samples. Buildings are clustered in the black area. For machine learning, only the dark blue \textit{zone of interest} with an horizontal refinement up to \qty{5}{\m} is kept and up to a height of \qty{50}{\m} to reduce the mesh size.}
    \label{fig:meshing}
\end{figure}

We then create CFD meshes from the generated building arrangements. An outlook on the mesh configuration is shown in Figure \ref{fig:meshing}. The domain is rectangular, with the buildings located in the upwind part, in order to have long areas for long distance wakes to develop. Additionally, the mesh is such that a restricted area including buildings and wakes could be chosen in the postprocessing step to reduce the mesh size for machine learning applications. Targeted horizontal refinement was $\qty{2}{\m}$ near the buildings, $\qty{5}{\m}$ in the area of interest, and $\qty{10}{\m}$ in the rest of the mesh. Vertical refinement varies from $\qty{1}{\m}$ near the ground to $\qty{50}{\m}$ at the top of the domain.

Simulations were then carried out using code\_saturne 8.0 \citep{Saturne2025}. Universal functions for atmospheric profiles are imposed as inlets on the upwind, side and top boundaries of the mesh. \citet{Hoegstroem1988} universal functions are used for the unstable stratifications, and \citet{Chenge2005} functions are used for stable stratifications. Logarithmic profiles are used in the neural stratification configuration. The downwind side of the mesh is left as a free outlet.

Additionally, two-scale log law wall functions with rugosity $z_0$ and temperature computed from the value of the potential temperature profile at ground level are imposed as boundary conditions on the ground.
Finally, wall functions with a constant roughness $z_\mathrm{buildings}=\qty{0.01}{\m}$ and no thermal fluxes are imposed as boundary conditions on the walls and roofs of the buildings.

For all simulations, a constant wind velocity of \qty{6}{\m\per\s} at a reference height of $\qty{80}{\m}$ is set. Additionally, a constant reference temperature at ground level of $\qty{20}{\celsius}$ is used to compute the profiles.

We use the \textit{dry atmosphere} model of code\_saturne, and a $k-\epsilon$ turbulence model with a linear turbulent production term. Cells are initialised with the values of the Monin-Obukhov profiles. We then run the simulations for $25,000$ iterations with a time step of $\qty{1}{s}$, until we reach a steady state. We average the last $1,500$ iterations to avoid selecting a particular time from a low-frequency oscillation such as Von-Karman vortex streets, resulting in a time-averaged steady state.

Final simulation outputs are atmospheric flow quantities defined at each cell center: the velocity field $\mathbf{v}$, the potential temperature $\theta$, the pressure variation $p$, the turbulent kinetic energy $k$ and the turbulent kinetic energy dissipation rate $\epsilon$.

\subsection{Postprocessing}

Finally, we postprocess the results for machine learning applications. As graph-based models require significant amounts of VRAM to train on large meshes, only the zone of interest, up to a height of $\qty{50}{\m}$, is kept for machine learning purposes (dark blue area in Figure \ref{fig:meshing}). This results in samples comprising at most $200k$ cells, which is within the range of sizes GNN models can handle without the partitioning of high-end GPU cards \citep{Liu2023}.

\section{Benchmark model parameters}\label{secA:params}

This appendix aims to provide more details on each model's hyperparameters choice for the benchmark. We provide elements at the origin of the code implementation and important hyperparameter choices below.

Hyperparameters were chosen to reach around $6M$ trainable parameters for each model, except for BSMGN, which was reduced to $0.4M$ parameters so that the training could be done on a single A100 GPU like the other models. Additionally, when possible, hyperparameters were chosen to match parameter choices made for the Drivaernet++ dataset in \citet{Alkin2025}, which is the one most resembling to our dataset among those studied in the different papers.

Finally, the training script was done using the tools of PhysicsNeMo \citep{NVIDIA2025} for optimization and profiling purposes.

\paragraph{AB-UPT}

We use AB-UPT's publicly available implementation, available at https://github.com/Emmi-AI/anchored-branched-universal-physics-transformers. However, since this implementation does not allow for extra features in the geometry point cloud, we modify the supernode pooling layer accordingly, and provide the modified code below. Additionally, since we do not use surfacic predictions in this study, no loss is computed on the surfacic branch's results, and the number of layers of the surfacic branch decoder is set at the minimum of $1$. Finally, the numbers of geometry points, supernodes, and surface anchor points were chosen to match the summed values of the terrain and obstacles' points and supernodes chosen for AB-SWIFT. The number of volume anchor points was kept the same as for AB-SWIFT.

\begin{table}[h]
    \centering
    \begin{tabular}{m{8.5cm}m{3.5cm}} \toprule
         Description & Value \\ \midrule
         Number of input points for the geometry, randomly subsampled from mesh points adjacent to the ground and buildings & $\num{8192}$ \\ 
         Number of supernodes for the geometry & $\num{2048}$ \\ 
         Number of anchor points in the surface point cloud, randomly subsampled from mesh points adjacent to the ground and buildings & $\num{2048}$ \\ 
         Hidden dimension & $192$ \\ 
         Number of anchor points in the volume point cloud, randomly subsampled from mesh cell centers & $\num{8192}$ \\ 
         Number of attention heads & $3$ \\ 
         Number of geometry transformer blocks & $1$ \\ 
         Number of physics blocks & $3$ \\ 
         Number of transformer layers in the volume decoder & $5$ \\ 
         Radius for supernodes pooling & $1$ \\ 
         Loss computation & On anchor points only \\ 
         \bottomrule
    \end{tabular}
    \caption{AB-UPT's chosen hyperparameters values}
\end{table}

\begin{minted}[%
 breaklines,
 frame=single,
 baselinestretch=0.8
 ]{python}
class SupernodePoolingRelPos(SupernodePoolingPosonly):
    '''Modification of AB-UPT's supernode pooling layer that takes into account features in addition of coordinates
    Only the relative positions mode is implemented.

    Args:
        radius: Radius around each supernode. From points within this radius, messages are passed to the supernode.
        k: Numer of neighbors for each supernode. From the k-NN points, messages are passed to the supernode.
        hidden_dim: Hidden dimension for positional embeddings, messages and the resulting output vector.
        ndim: Number of positional dimension (e.g., ndim=2 for a 2D position, ndim=3 for a 3D position)
        nfeat: Number of features
        max_degree: Maximum degree of the radius graph. Defaults to 32.'''
    def __init__(self,
                    hidden_dim: int,
                    ndim: int,
                    nfeat: int,
                    radius: float | None = None,
                    k: int | None = None,
                    max_degree: int = 32):
        super().__init__(hidden_dim, ndim, radius, k, max_degree, mode = 'relpos')

        #modify message dim to account for features
        message_input_dim = hidden_dim + nfeat
        
        self.message = nn.Sequential(
            nn.Linear(message_input_dim, hidden_dim),
            nn.GELU(),
            nn.Linear(hidden_dim, hidden_dim),
        )
        #also modify projection
        self.proj = nn.Linear( 2 * hidden_dim + nfeat, hidden_dim)

    def create_messages(self, input_pos, src_idx, dst_idx, supernode_idx):
        '''Create messages with features as well as positions'''

        #split positions and features
        input_pos, input_feat = input_pos[... ,:self.ndim], input_pos[..., self.ndim:]
        
        #embed positions
        src_pos = input_pos[src_idx]
        dst_pos = input_pos[dst_idx]
        dist = dst_pos - src_pos
        mag = dist.norm(dim=1).unsqueeze(-1)
        x_pos = self.rel_pos_embed(torch.concat([dist, mag], dim=1))

        #concatenate positions and features
        x = torch.concat([x_pos, input_feat[src_idx]], dim = -1)
        
        supernode_feat = input_feat[supernode_idx]
        supernode_pos_embed = self.pos_embed(input_pos[supernode_idx])
        supernode_embed = torch.concat([supernode_pos_embed, supernode_feat], dim = -1)

        #message
        x = self.message(x)

        return x, supernode_embed
\end{minted}

\paragraph{GAOT}

We use GAOT's publicly available implementation (https://github.com/camlab-ethz/GAOT), without modification. The number of latent tokens is chosen to match AB-SWIFT and AB-UPT's numbers of anchor points.

\begin{table}[h]
    \centering
    \begin{tabular}{m{8.5cm}m{3.5cm}} \toprule
         Description & Value \\ \midrule
         Number of latent tokens in $x$, $y$, and $z$ dimensions & $32,32,8$ \\ 
         Multiscale radii in the MAGNO encoder and decoder & $1, 5, 10, 20, 50$ \\ 
         Learnable scale weights in the MAGNO encoder and decoder & True \\ 
         Number of MLP layers at the end of the MAGNO encoder & $\num{3}$ \\ 
         Hidden size of the MAGNO encoder and decoder & $\num{64}$ \\ 
         Hidden size of MAGNO's output & $128$ \\ 
         Edge sample ratio & $0.7$ \\ 
         Vision transformer's patch size & $2$ \\ 
         Vision transformer's hidden size & $128$ \\ 
         Number of transformer layers & $8$ \\ 
         Number of attention heads & $8$ \\ 
         Positional encoding & Rotary positional encoding \\
         \bottomrule
    \end{tabular}
    \caption{GAOT's chosen hyperparameters values.}
\end{table}

\paragraph{Transolver}

We use PhysicsNeMo's Transolver implementation, without modification (https://github.com/NVIDIA/physicsnemo/tree/main/physicsnemo/models/transolver).

\begin{table}[h]
    \centering
    \begin{tabular}{m{8.5cm}m{3.5cm}} \toprule
         Description & Value \\ \midrule
         Number of physics attention layers & $8$ \\ 
         Number of physics attention slices & $64$ \\ 
         Hidden size & $256$ \\ 
         Number of attention heads & $8$ \\ 
         Activation function & GeLU \\
         \bottomrule
    \end{tabular}
    \caption{Transolver's chosen hyperparameters values.}
\end{table}

\paragraph{BSMGN}

We use PhysicsNeMo's BSMGN implementation, without modification.

\begin{table}[h]
    \centering
    \begin{tabular}{m{8.5cm}m{3.5cm}} \toprule
         Description & Value \\ \midrule
         single-scale MGN processor size & $0$ (i.e. not used)\\ 
         Number of u-net like mesh levels & $5$ \\ 
         Hidden size & $64$ \\ 
         Number of hidden layers per message passing & 2 \\ 
         Number of calls to the bistride processor & 1 \\
         \bottomrule
    \end{tabular}
    \caption{BSMGN's chosen hyperparameters values.}
\end{table}


\printcredits

\section*{Statements and Declarations}

\begin{itemize}
\item Funding 

This work was funded by the French National Association of Research and Technology (ANRT), EDF R\&D and the CEA DASE with the Industrial Conventions for Training through REsearch (CIFRE grant agreement 2023/1614). The authors acknowledge their support, as well as SINCLAIR AI lab for helpful discussions.

\item Competing interests

The authors have no relevant financial or non-financial interests to disclose.

\item Code and Data availability

The code and data developed for the current study is available on github at https://github.com/cerea-daml/abswift.

\end{itemize}

\bibliographystyle{cas-model2-names}

\bibliography{abswift_paper}

\end{document}